\documentclass[letterpaper, 10 pt, conference]{ieeeconf}  

\IEEEoverridecommandlockouts                              

\usepackage{graphicx}
\usepackage{epsfig}
\usepackage{epic,eepic}
\usepackage{times}
\usepackage{mathptmx}
\usepackage{amsfonts}
\usepackage{amsmath}
\usepackage{cite}

\usepackage[dvipdfm]{color}

\usepackage{times}
\usepackage{multirow}

\definecolor{red}{rgb}{1,0,0}
\definecolor{green}{rgb}{0,1,0}
\definecolor{blue}{rgb}{0,0,1}
\definecolor{violet}{rgb}{1,0,1}
\definecolor{cyan}{cmyk}{1,0,0,0}
\definecolor{magenta}{cmyk}{0,1,0,0}
\definecolor{yellow}{cmyk}{0,0,1,0}
\definecolor{white}{rgb}{1,1,1}

\newcommand{\CommentOut}[1]{}

\newcommand{\FIG}[3]{
\begin{minipage}[b]{#1cm}
\begin{center}
\includegraphics[width=#1cm]{#2}
{\scriptsize #3}
\end{center}
\end{minipage}
}

\newcommand{\FIGR}[3]{
\begin{minipage}[b]{#1cm}
\begin{center}
\includegraphics[angle=-90,clip,width=#1cm]{#2}\vspace*{1mm}
\\
{\scriptsize #3}
\vspace*{1mm}
\end{center}
\end{minipage}
}

\begin{document}

\thispagestyle{empty}
\pagestyle{empty}

\title{\LARGE \bf
Self-localization from Images with Small Overlap
}

\author{Tanaka Kanji
\thanks{Our work has been supported in part by 
JSPS KAKENHI 
Grant-in-Aid for Young Scientists (B) 23700229,
and for Scientific Research (C) 26330297.}
\thanks{K. Tanaka is with Faculty of Engineering, University of Fukui, Japan.
{\tt\small tnkknj@u-fukui.ac.jp}}
\thanks{We would like to express our sincere
graditude to Kentaro Yanagihara and Atsushi Yoshikawa for initial investigation on scene recognition tasks on the dataset, which helped us to focus on our PCA-NBNN project. }
}

\maketitle

\begin{abstract}
With the recent success of visual features from deep convolutional
neural networks (DCNN) in visual robot self-localization,
it has become important and practical to address more general 
self-localization scenarios. In this paper, we address the scenario of
self-localization from images with small overlap. We explicitly
introduce a localization difficulty index as a decreasing function
of view overlap between query and relevant database images and
investigate performance versus difficulty for 
challenging cross-view self-localization tasks.
We then reformulate the self-localization as a scalable bag-of-visual-features (BoVF) scene retrieval and present an efficient
solution called PCA-NBNN, aiming to facilitate fast and yet 
discriminative correspondence between partially overlapping images.
The proposed approach adopts recent findings in discriminativity preserving
encoding of DCNN features using principal component analysis
(PCA) and cross-domain scene matching using naive Bayes nearest
neighbor distance metric (NBNN). We experimentally demonstrate that
the proposed PCA-NBNN framework frequently achieves comparable results
to previous DCNN features and that the BoVF model
is significantly more efficient. We further address an important 
alternative scenario of ``self-localization from images with NO
overlap" and report the result.
\end{abstract}

\newcommand{\figU}{
\begin{figure*}[t]
\begin{center}
\FIG{17}{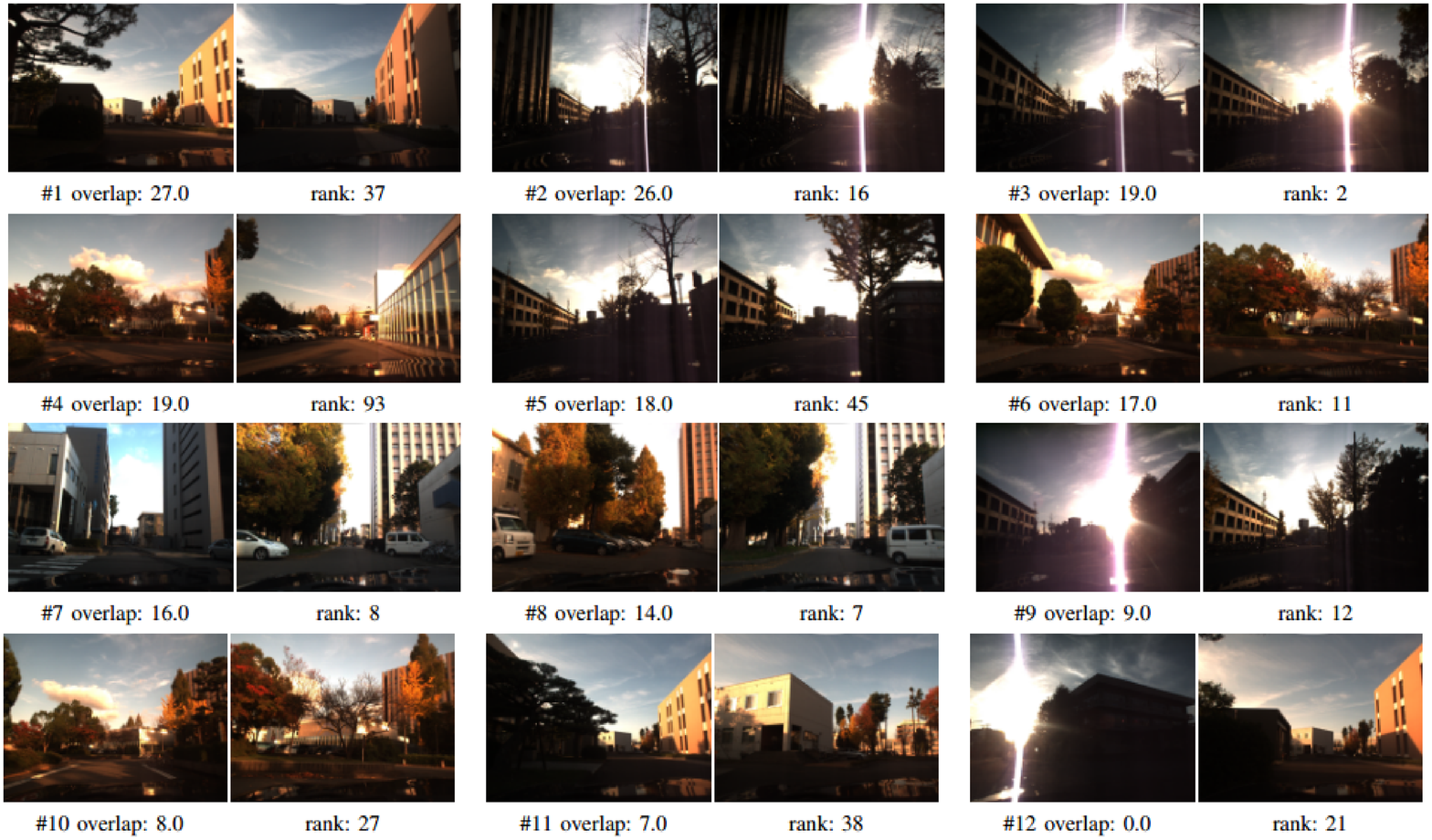}{}
\caption{Samples of self-localization tasks. 
Displayed in figures are samples of self-localization tasks (using ``bodw20" algorithm). 
We uniformly sampled them from the experiments. 
For each sample, its query image (left) and the relevant database image (right) are displayed with the view overlap score (``overlap") as well as the localization performance (``rank"). Here, ``rank" is the rank assigned to the ground-truth database relevant image, within a ranked list output by the recognition algorithm. From top to bottom, left to right, these samples are displayed in descending order of view overlap (i.e., from easiest to hardest).}\label{fig:U}
\vspace*{-5mm}
\end{center}
\end{figure*}
}

\newcommand{\figA}{
\begin{figure}[t]
\begin{center}
\begin{center}
\FIG{8.5}{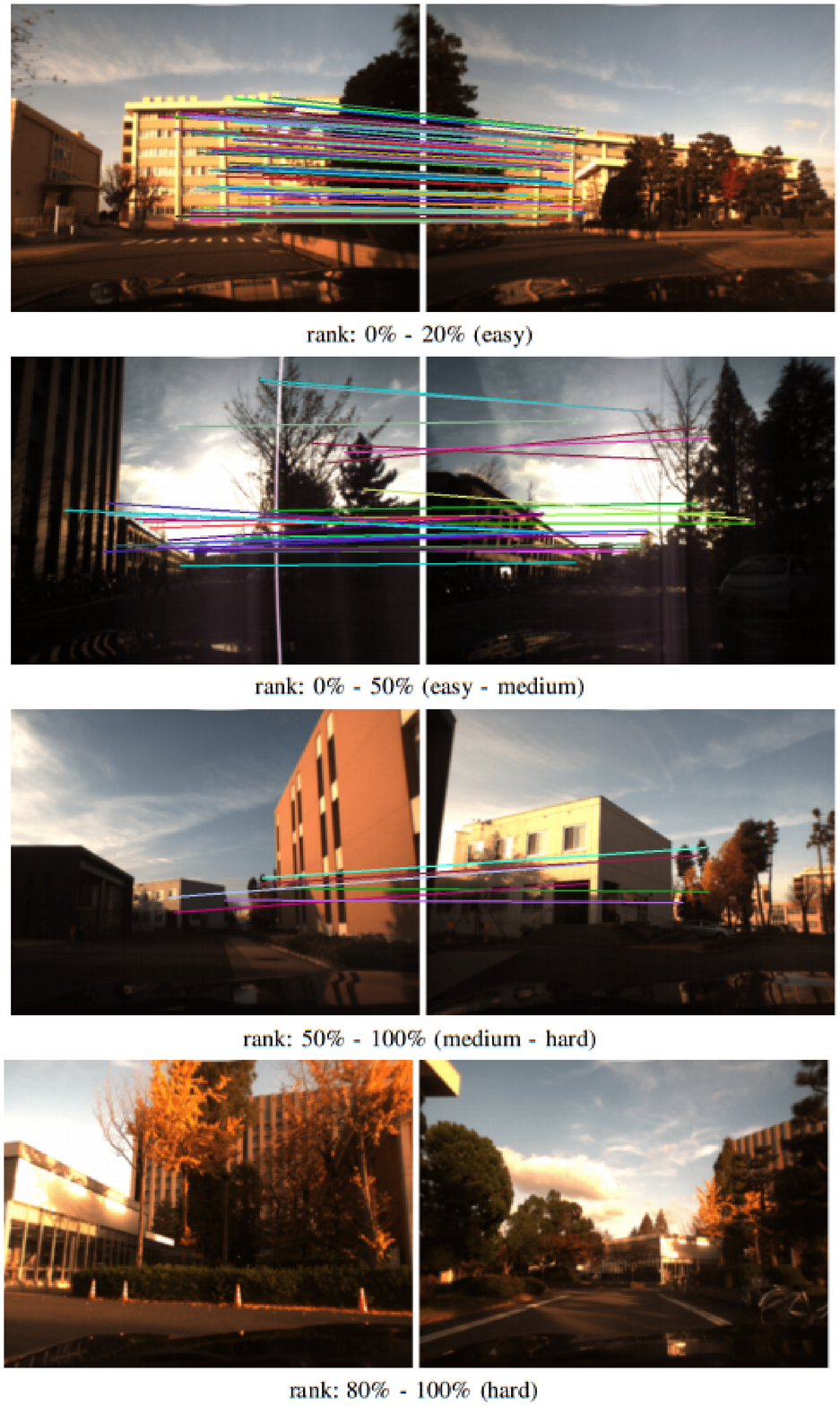}{}
\caption{Self-localization with different levels of localization difficulty index (LDI).
The LDI of a self-localization task
is a decreasing function of view overlap
between the query and relevant database image pair.
In experiments, 
we employ SIFT matching with VFC verification (colored line segments) to evaluate the amount of view overlap.
All the pairs in the dataset are 
evaluated and sorted according in ascending order of LDI.
Rank in the sorted list (normalized by the list's length) [\%]
can be viewed as a prediction of
relative difficulty of 
the corresponding self-localization task.
Displayed in figures
are samples from 
self-localization tasks
with four different levels of ranks [\%].}\label{fig:A}
\end{center}
\end{center}
\end{figure}
}

\newcommand{\figB}[2]{
\begin{figure}[t]
\begin{center}
\begin{center}
\hspace*{-2mm}\FIG{8.5}{#1}{}
\caption{#2}\label{fig:B}
\end{center}
\end{center}
\end{figure}
}

\newcommand{\figE}{
\begin{figure*}[t]
\begin{center}
\begin{center}
\FIGR{6}{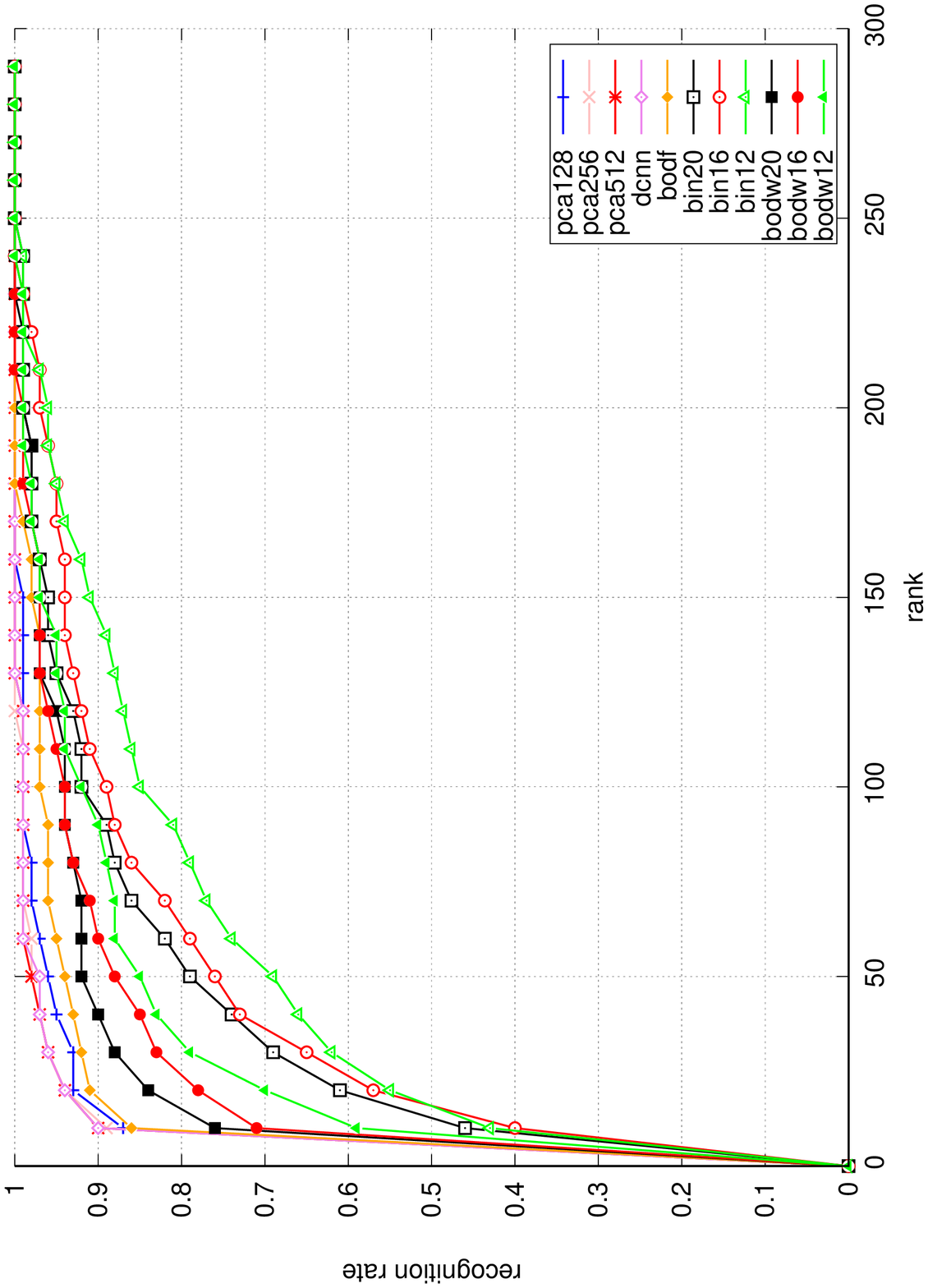}{rank: 0\% - 20\% }\hspace*{-5mm}\FIGR{6}{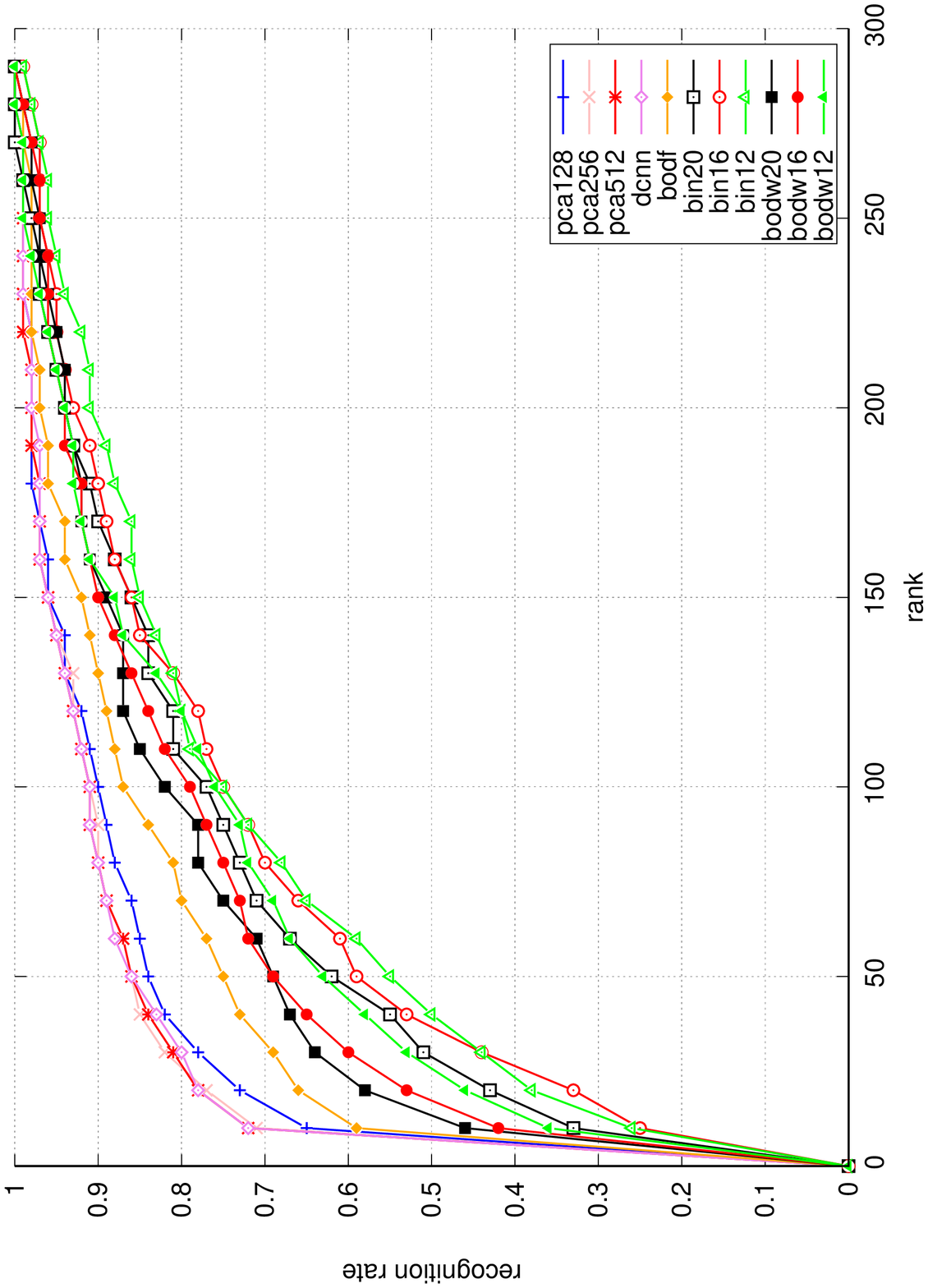}{rank: 0\% - 50\%}\hspace*{-5mm}\FIGR{6}{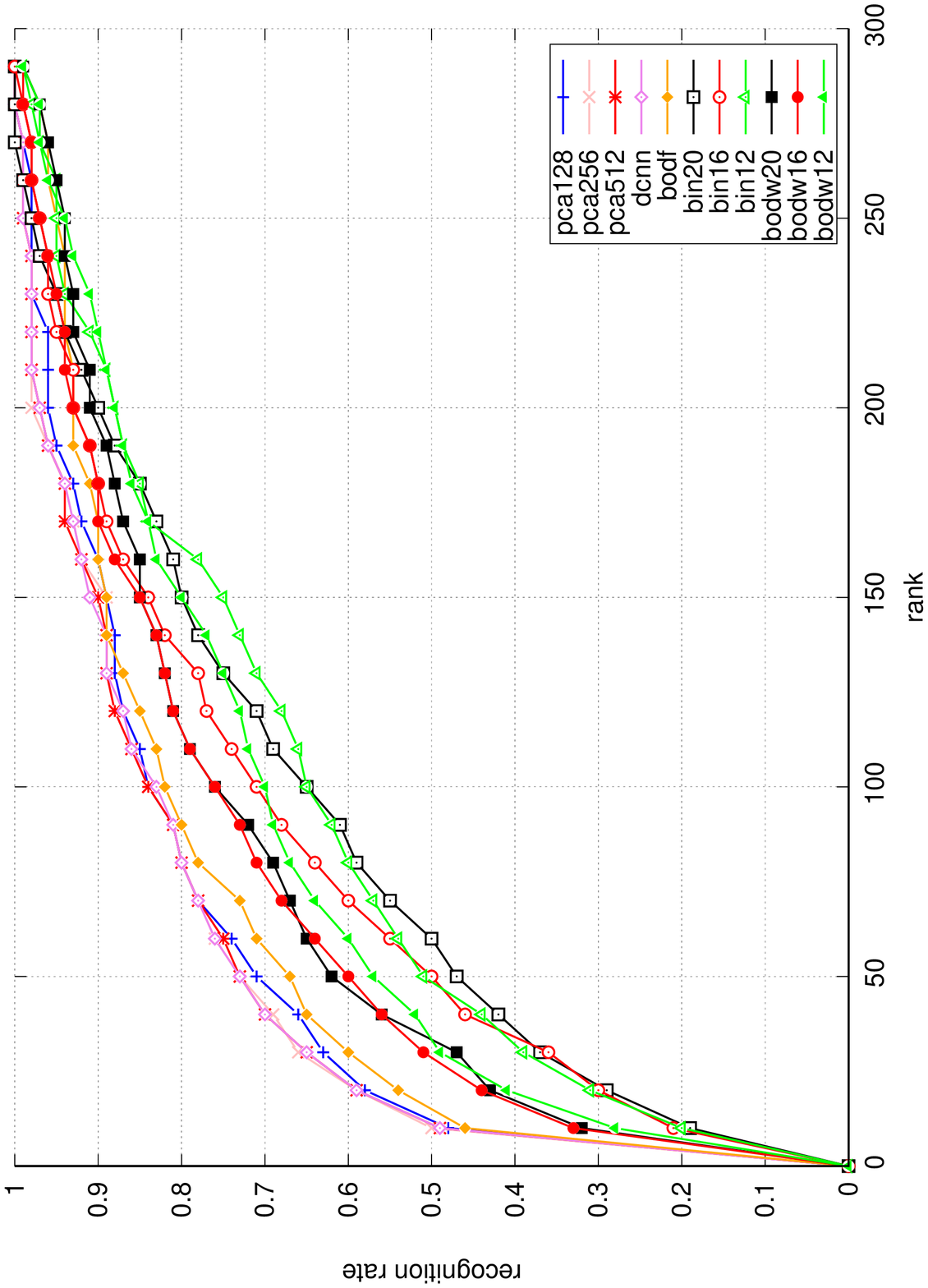}{rank: 0\% - 100\%}\vspace*{-3mm}\\
\caption{Localization performance on relatively easy localization scenarios.}\label{fig:E}\vspace*{-7mm}
\end{center}
\end{center}
\end{figure*}
}

\newcommand{\figH}{
\begin{figure*}[t]
\begin{center}
\begin{center}
\FIGR{6}{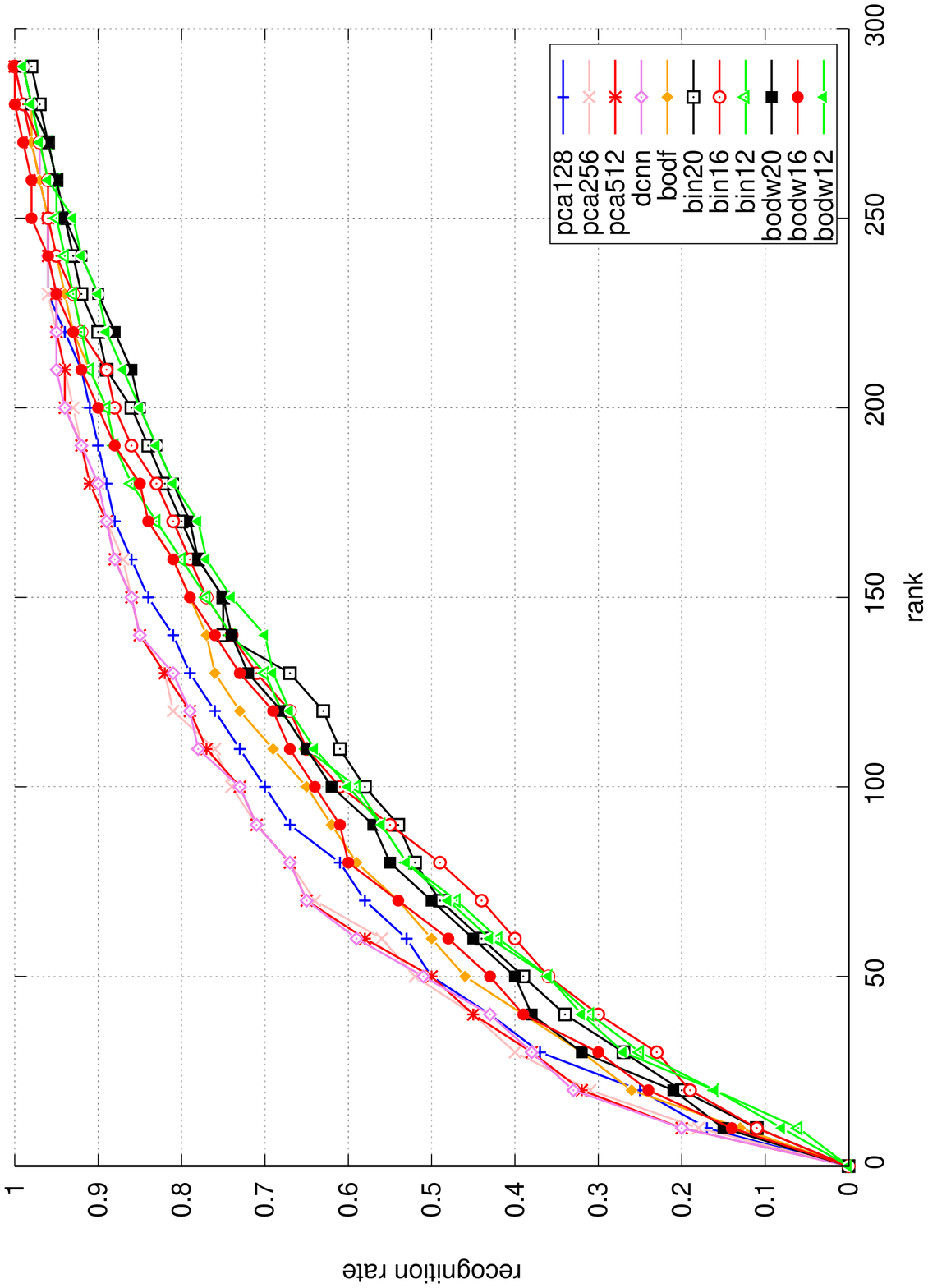}{rank: 50\% - 100\%}\hspace*{-5mm}\FIGR{6}{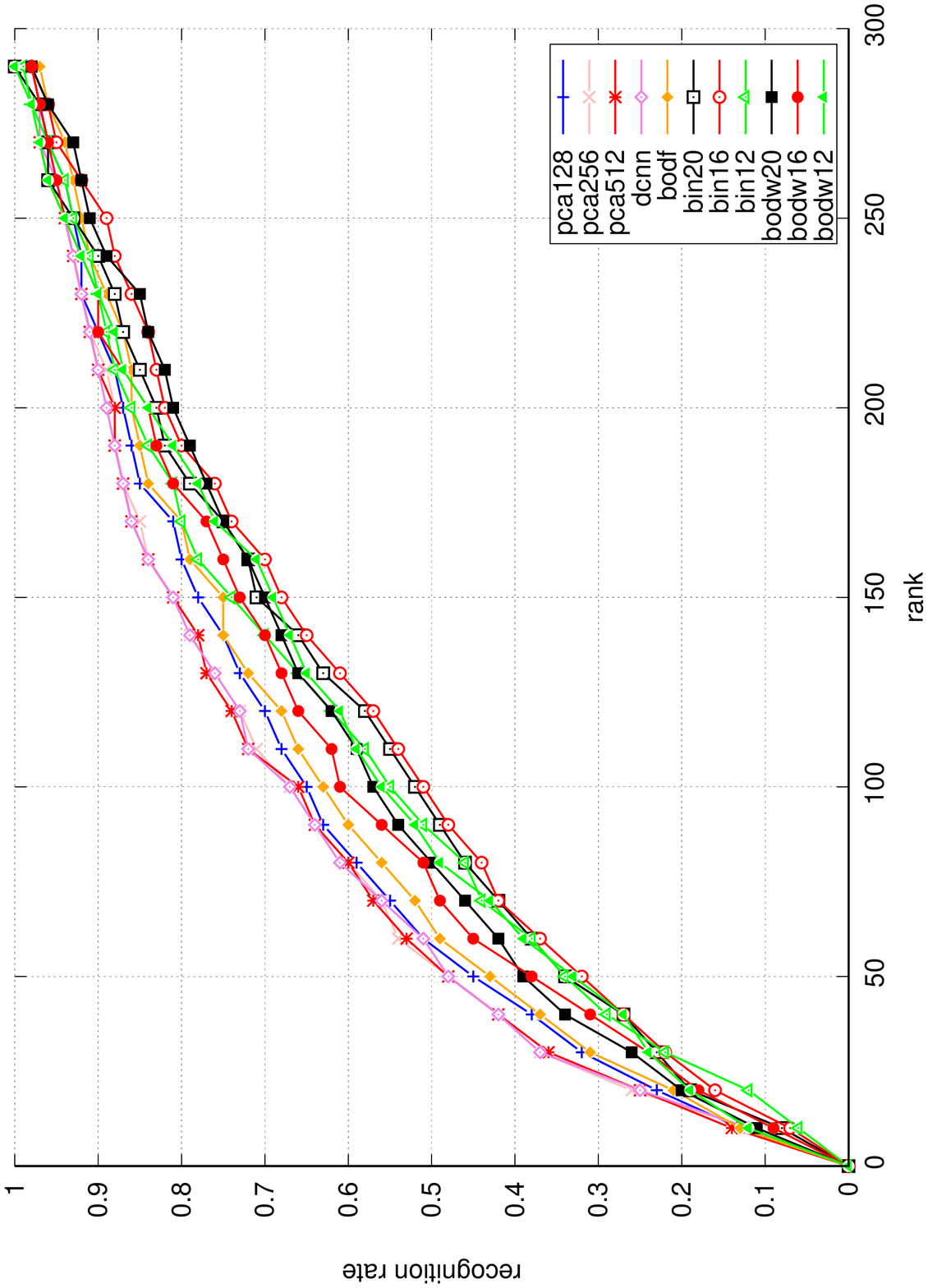}{rank: 80\% - 100\%}\hspace*{-5mm}\FIGR{6}{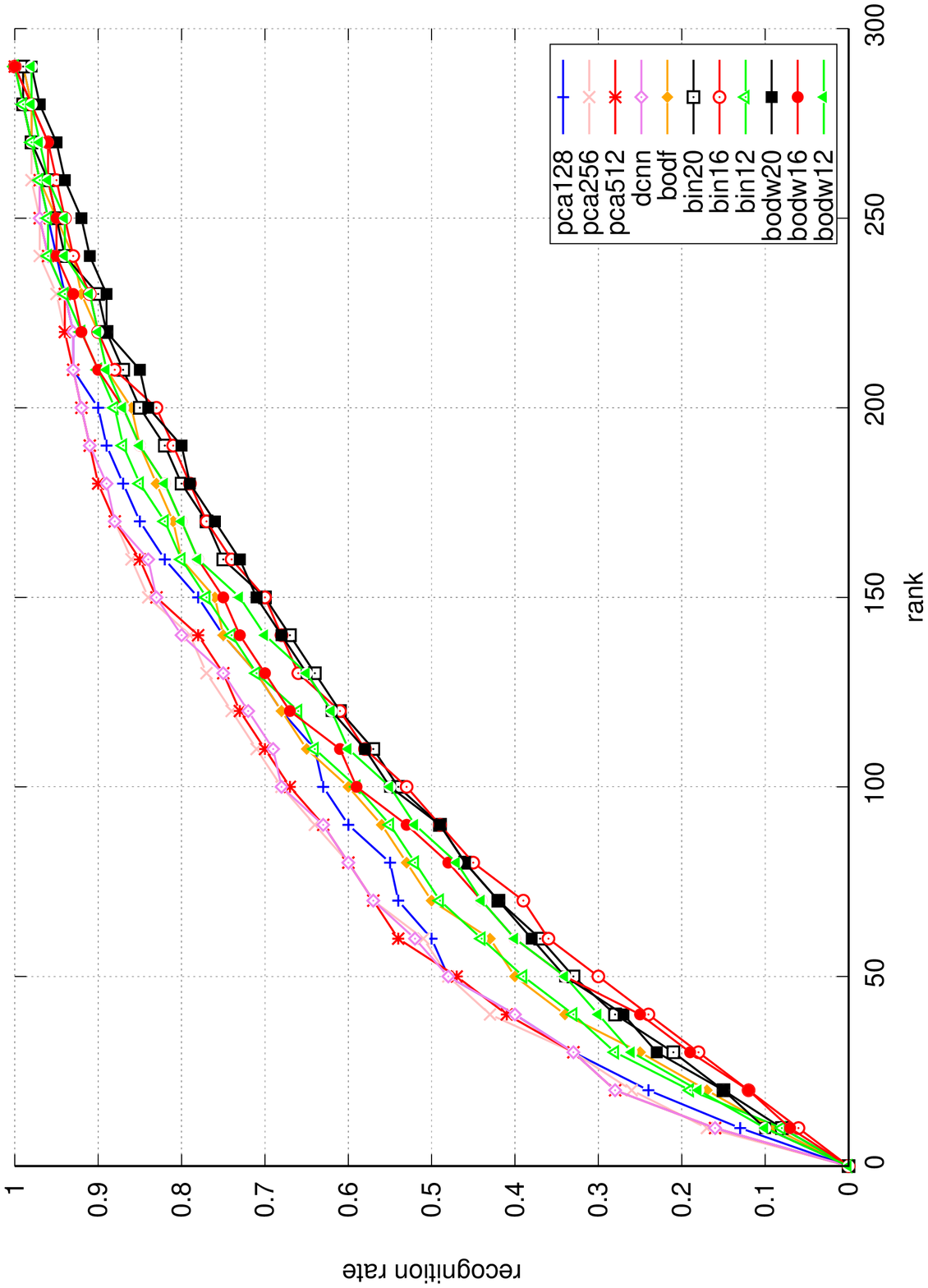}{NO}\vspace*{-3mm}%
\caption{Localization performance on relatively hard localization scenarios.}\label{fig:H}\vspace*{-7mm}
\end{center}
\end{center}
\end{figure*}
}
\newcommand{\figBa}{\figB{res2_all_top20.eps}{top 20\%}}
\newcommand{\figBb}{\figB{res2_all_top.eps}{top 50\%}}
\newcommand{\figBc}{\figB{res2_all_random.eps}{random}}
\newcommand{\figBd}{\figB{res2_all_bottom.eps}{bottom 50\%}}
\newcommand{\figBe}{\figB{res2_all_bottom20.eps}{bottom 20\%}}
\newcommand{\figBf}{\figB{res2_all_no.eps}{no overlap}}

\newcommand{\figC}[2]{
\begin{figure}[t]
\begin{center}
\begin{center}
\hspace*{-2mm}\FIG{8.5}{#1}{}
\caption{#2}\label{fig:C}
\end{center}
\end{center}
\end{figure}
}

\newcommand{\figCa}{\figC{cvl/fig3/exp4/top20/marrows/38.eps}{}}
\newcommand{\figCb}{\figC{cvl/fig3/exp4/top/marrows/48.eps}{}}
\newcommand{\figCc}{\figC{cvl/fig3/exp4/bottom/marrows/88.eps}{}}
\newcommand{\figCd}{\figC{cvl/fig3/exp4/bottom20/marrows/58.eps}{}}

\newcommand{\figD}{
\begin{figure*}[t]
\begin{center}
\begin{center}
\FIG{6}{cvl/fig3/exp4/top20/marrows/38.eps}{}\hspace*{-5mm}%
\FIG{6}{cvl/fig3/exp6/bottom/marrows/88.eps}{}\hspace*{-5mm}%
\FIG{6}{cvl/fig3/exp5/bottom20/marrows/28.eps}{}
\caption{figD}\label{fig:D}
\end{center}
\end{center}
\end{figure*}
}

\renewcommand{\figD}{
\begin{figure*}[t]
\begin{center}
\begin{center}
\FIG{17}{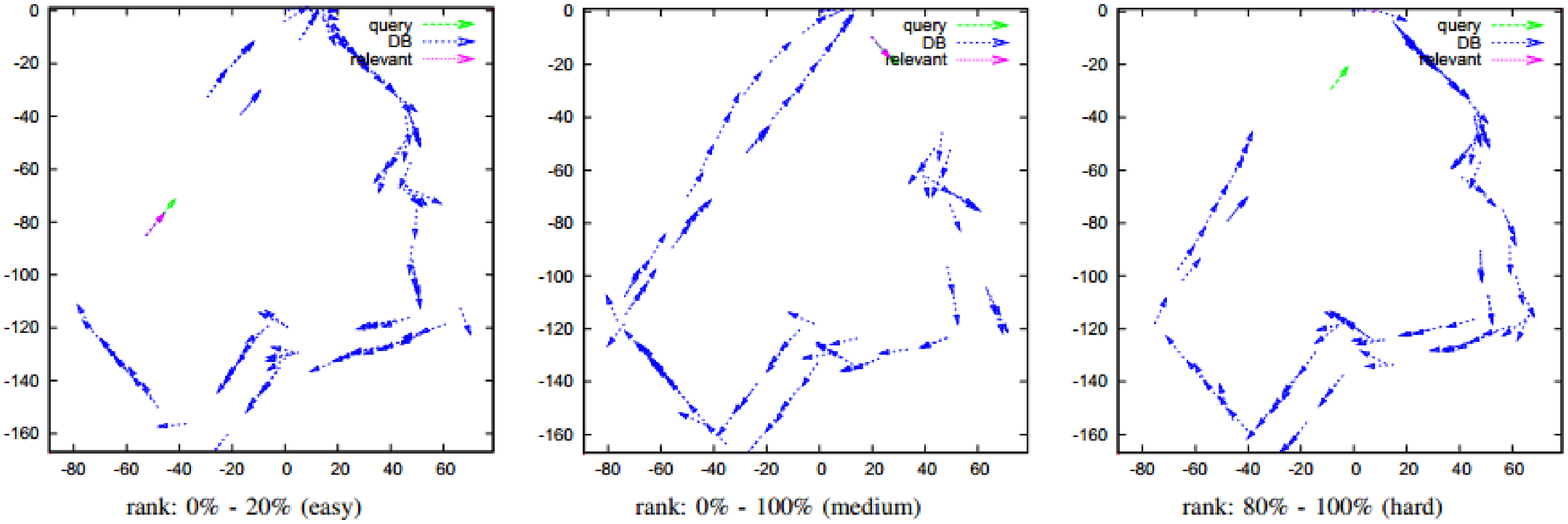}{}%
\caption{Sample configurations of viewpoints for different levels of localization difficulties.}\label{fig:D}
\end{center}
\vspace*{-5mm}
\end{center}
\end{figure*}
}

\newcommand{\figF}{
\begin{figure}[t]
\begin{center}
\begin{center}
\FIG{8.5}{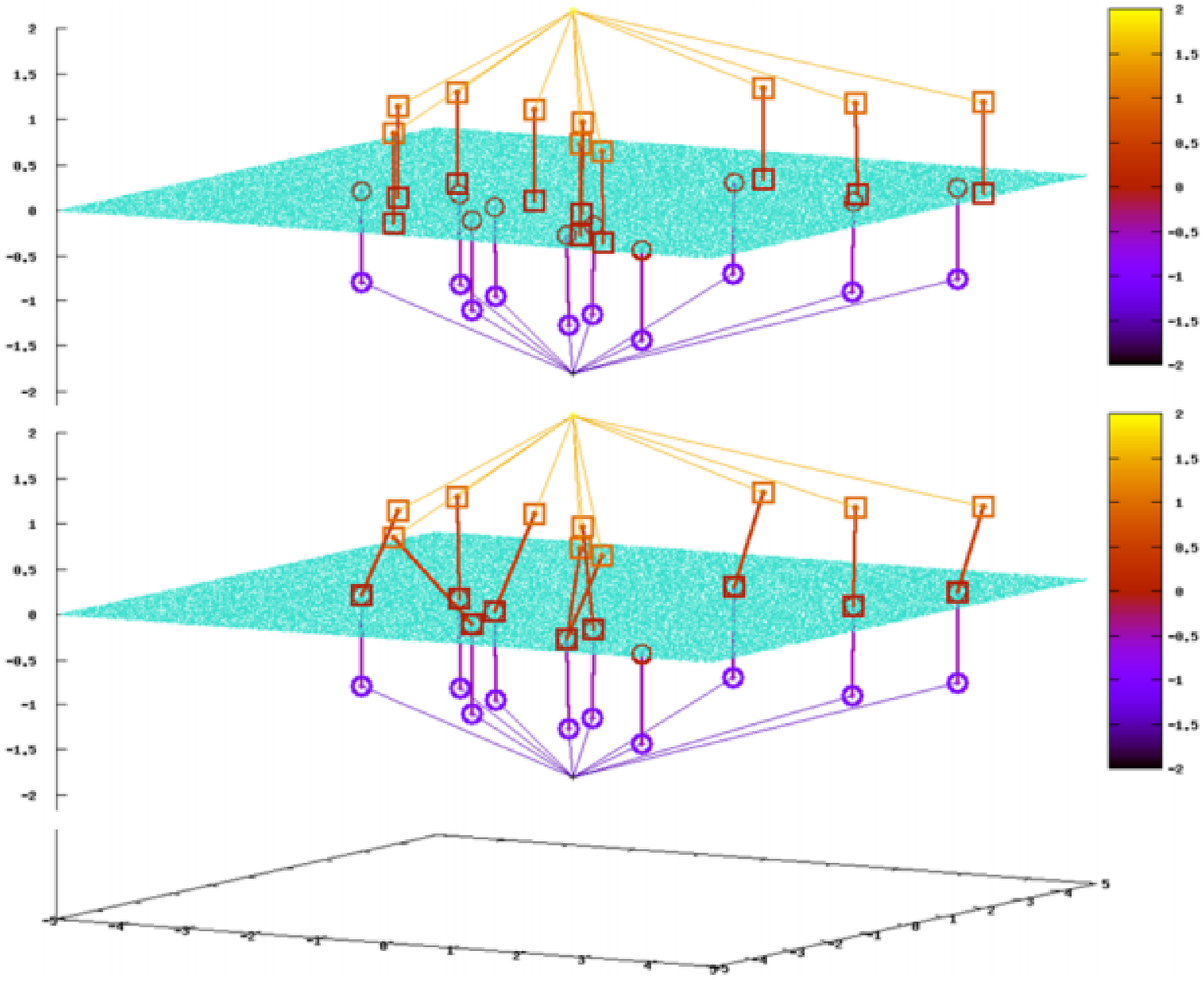}{}
\caption{Effect of asymmetric distance computation (ADC). 
The figures 
compare
the two different encoding schemes,
BoW (top) and ADC (bottom),
using a toy example of a 2D feature space $x$-$y$,
in the case of the fine library.
In the figures,
query/database images are located $z=2$/$z=-2$,
local features extracted from query/database images are located $z=1$/$z=-1$,
and
library features (green dots) including NN library features (colored small boxes) are located $z=0$.
Previous BoW systems (top),
which encode both query and database features,
frequently fail to identify common library features
between query and database images in the case of our fine library.
Conversely,
ADC, which encodes only database features, not query features, 
is stable to identify NN library features of individual database features by an 
online search over the space of library features (i.e., $z=0$).}\label{fig:F}
\end{center}
\vspace*{-5mm}
\end{center}
\end{figure}
}

\newcommand{\figG}{
\begin{figure*}[t]
\begin{center}
\begin{center}
\FIG{17}{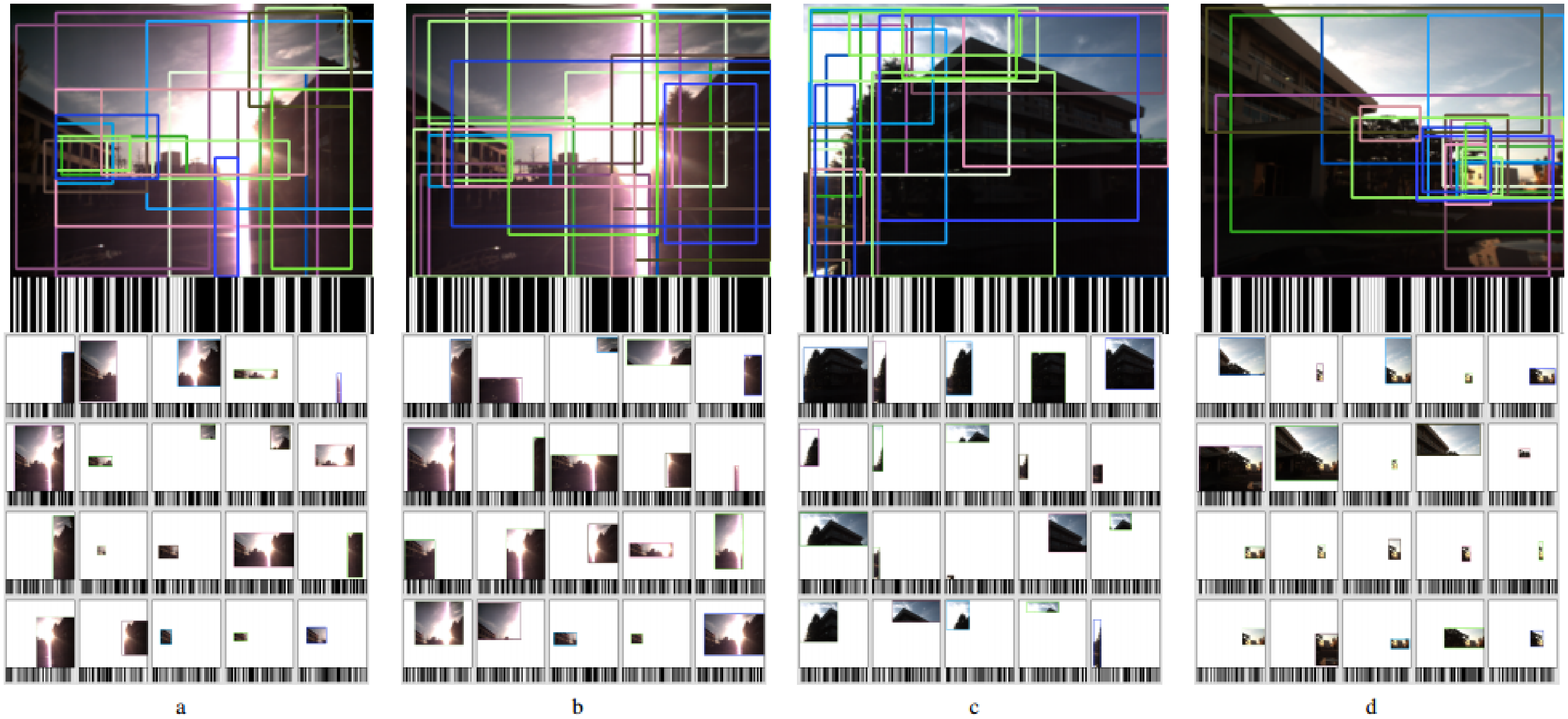}{}
\caption{Compact binary landmarks. 
a, b, c, and d:
4 different examples of
a query image (top)
being explained by
one image-level feature and
20 part-level features (bottom).
Each scene part is further encoded to a 128-bit binary code,
which is visualized by a barcode.}\label{fig:G}
\end{center}
\vspace*{-5mm}
\end{center}
\end{figure*}
}

\newcommand{\figI}{
\begin{figure}[t]
\begin{center}
\begin{center}
\FIG{8.5}{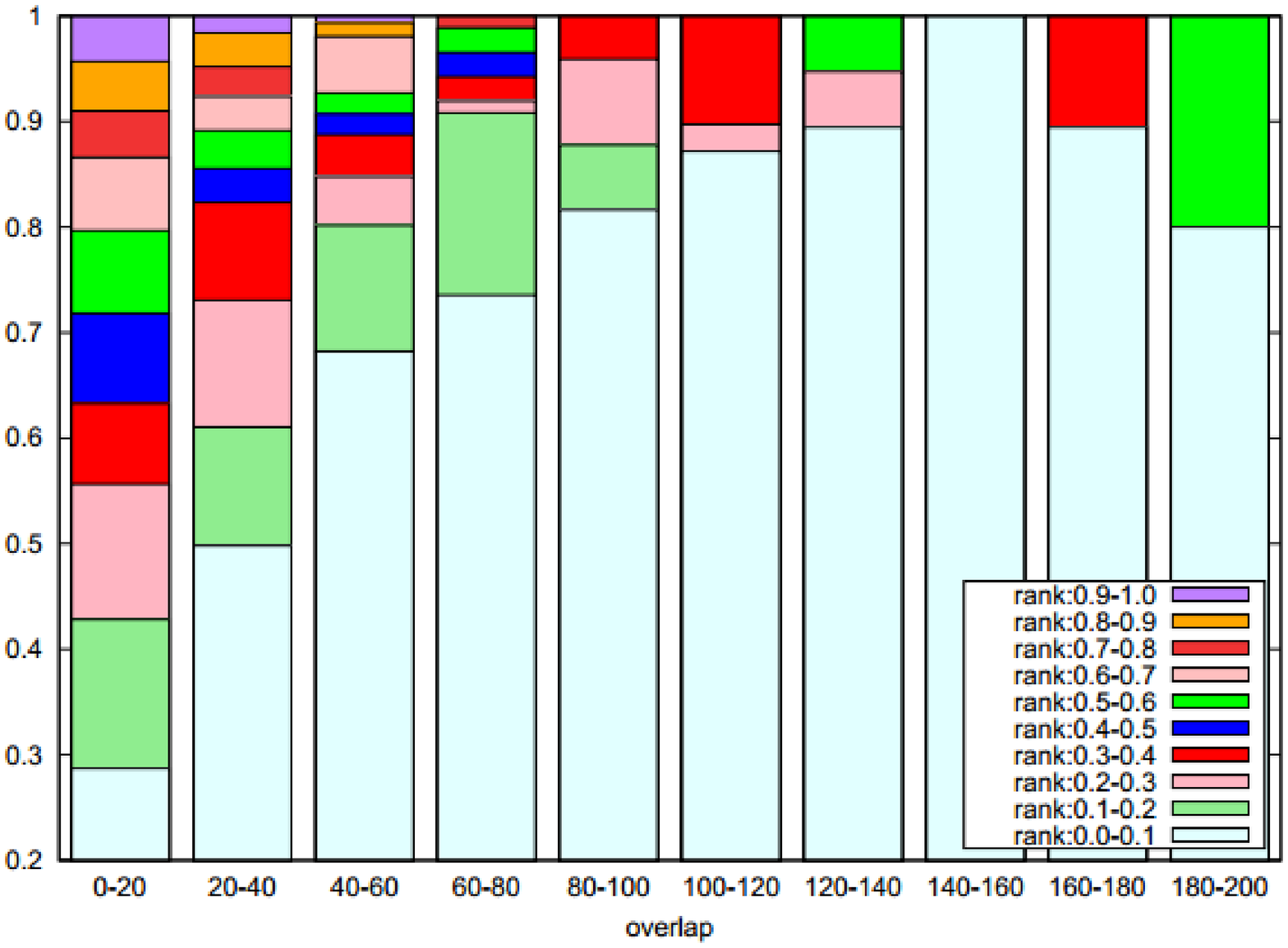}{}
\vspace*{-7mm}
\caption{Performance vs. difficulty. 
Vertical axis:
ratio of 
self-localization tasks
where the ground truth image pair is top-$X$ ranked
for ten different ranges of rank $X$ 
($X$: 0.0-0.1, 0.1-0.2, ... , 0.9-1.0.).
Horizontal axis: 
view overlap in terms of number of VFC matches,
which is a decreasing function of localization difficulty index.}\label{fig:I}
\end{center}
\vspace*{-5mm}
\end{center}
\end{figure}
}

\newcommand{\figL}{
\begin{figure}[t]
\begin{center}
\begin{center}
\FIG{8}{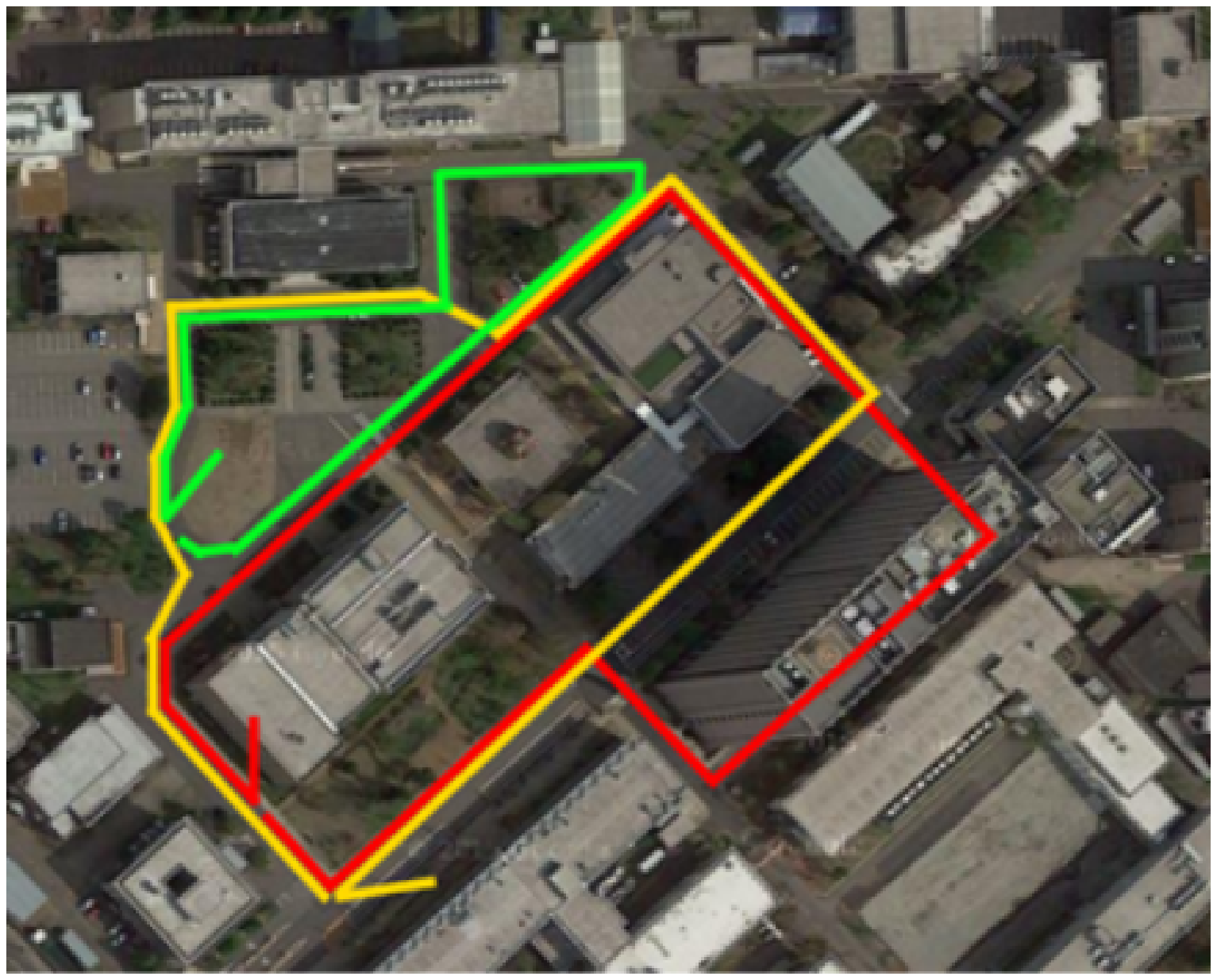}{}
\caption{Experimental environments. Red, yellow, and green lines: 
viewpoint paths on which dataset \#1, \#2, and \#3 were collected.}\label{fig:L}
\end{center}
\vspace*{-5mm}
\end{center}
\end{figure}
}

\figA

\section{Introduction}

With the recent success of visual features from deep convolutional neural networks (DCNN) in visual robot self-localization, it has become important and practical to address more general self-localization  scenarios. Self-localization aims to use a robot's visual image as a query input and to search over a database of pre-mapped images to locate a relevant database image that is viewed from the nearest neighbor viewpoint to the query image's viewpoint. Recently, it has been found that the intermediate responses of a DCNN can be viewed as a discriminative feature for image matching. In \cite{neuralcodes}, the DCNN descriptor is exploited for the image retrieval task where DCNN descriptors are translated to short vectors by PCA dimension reduction. In \cite{iros15dcnn}, DCNN descriptors are applied to visual robot self-localization tasks and produce impressive results.

In this paper, we address the problem of self-localization from images
with small view overlap. This is a challenging scenario with important
applications including self-localization using far features \cite{farfeatures}, object co-segmentation \cite{coseg}, sparse feature maps \cite{contreras2015trajectory}, and the lost robot problem \cite{kidnapped}. To date, the majority of the existing work on self-localization, including those with DCNN features, rely on a strong assumption: there is a large view overlap (e.g., $>50\%$) between the query and relevant database images. In general
cases where the overlap between two relevant views is frequently small (e.g., $\le 10$\%), the self-localization problem is largely unsolved.

To address the above issue, we explicitly introduce a localization
difficulty index as a decreasing function of view overlap between
the query and relevant database images (Fig. \ref{fig:A}), and investigate the performance versus difficulty of challenging cross-view self-localization tasks. We collected a dataset of view images with ground truth viewpoints, and evaluated amount of view overlap for each relevant image pair by employing techniques of
common visual pattern discovery \cite{vfc}. 
We experimentally determined that
DCNN features fail in the case of small overlap owing to a
large number of outlier features and occlusions. We further address the
challenging and important scenario of ``self-localization from images
with NO overlap" and report the result.

We then reformulate the self-localization as a scalable bag-of-visual-features (BoVF) scene retrieval \cite{17} and present an efficient solution called PCA-NBNN, aiming to facilitate fast, yet discriminative correspondence between partially overlapping images. The basic idea is to encode local part-level DCNN features of a scene image into a BoVF document model and then apply an effective document retrieval technique for efficient indexing and search. Our encoding model adopts recent findings in discriminativity preserving encoding of DCNN features \cite{neuralcodes} where principal component analysis (PCA) compression provides efficient short codes that provide state-of-the-art accuracy on a number of recognition tasks. We also adopt a naive Bayes nearest neighbor distance metric (NBNN), inspired by our previous IROS15 paper, that has proven to be effective in an alternative application of cross-domain scene matching based on SIFT features \cite{kanji2015cross}. In experiments, we confirm that the proposed framework 
frequently achieves comparable results to previous DCNN features even though the BoVF model is significantly more efficient.

\subsection{Relation to Other Work}

The main contribution of this paper is in investigating the use of
DCNN features in challenging cross-view self-localization scenarios
and presenting an efficient recognition approach based on a BoVF scene
model. The BoVF subsystem employed in \ref{sec:partmodel} is inspired by a 
bag-of-parts model in the authors' previous ICRA15 paper \cite{kanji2015unsupervised}.

Scene descriptors for visual place recognition (VPR) problems have
been studied extensively. Local feature approaches such as BoVF
scene descriptors have been widely studied from various aspects \cite{arandjelovic2012three}
including confusing features, quantization errors, query expansion,
database augmentation, vocabulary tree, and global spatial geometric
verification as post-processing. As suggested by previous studies \cite{seqSLAM} and also by our ICRA15 paper \cite{icra15ando}, existing BoVF models are not sufficiently
discriminative and frequently fail to capture the appearance changes across
domains.

Global feature approaches such as the GIST feature descriptor
\cite{globaldescriptor1} (where a scene is represented by a single global feature
vector) are compact and have high matching speeds. In the robot
vision community, global feature approaches have been widely used
in the context of cross-domain VPR \cite{seqSLAM,dynamicslam4,dynamicslam3}. 
\cite{seqSLAM} introduces a
robust VPR framework called SeqSLAM for cross-season navigation
tasks separated by months or years and opposite seasons. More recently, in [2], the
authors demonstrated that DCNN features outperform the majority of the existing global features in typical VPR tasks.

In this study, the proposed approach is built on some of our previous
techniques including compact binary landmarks of deep network in
ICRA10 \cite{IkedaICRA10}, 
compact projection in IROS11 \cite{min2010compact}, 
NBNN scene descriptor in IROS15 \cite{kanji2015cross}, 
and bag-of-parts in ICRA15 \cite{kanji2015unsupervised}. 
However, the current study focuses on the use of DCNN features in visual robot
localization.

DCNN features have received considerable attention in the past years. 
However, effective use of DCNN descriptors in the context
of robot localization has not thus far been sufficiently explored and a main
topic of on-going research \cite{iros15dcnn}. In particular, the issue of view 
overlap as localization difficulty index
and the use of the PCA-NBNN model to address partially overlapping views have
not been addressed in existing studies.

\figL

\figD

\section{Problem}

\subsection{Dataset}\label{sec:dataset}

For clarity of presentation, we first describe the experimental
system by which a dataset is collected 
in our university campus (Fig. \ref{fig:L})
and used as a benchmark for
performance comparison in the experimental section. Although our
application scenario is single-view self-localization, we employed a
stereo SLAM system with visual odometry to collect a set of view
images with ground-truth viewpoint information. Our stereo SLAM
system is built on a Bumblebee stereo vision camera system and visual
odometry \cite{geiger2011stereoscan} and follows the standard formulation of pose graph SLAM \cite{ProbabilisticsRobotics}. We used images with size $640\times 480$ [pixels] from the
left eye view of the stereo camera as the image dataset.

\subsection{Localization Performance Index}\label{sec:lpi}

We conduct a series of $N^E=1,800$ self-localization tasks
using a set of $N^E$ independent subsets of the dataset.
For each task,
we sample 
one image $I^Q$ as a query input,
one image $I^R$ as a relevant image,
and a size $N^D-1$ 
image collection 
($N^D=100$)
as destructor images $\{I^D\}_{i=1}^{N^D-1}$
so that 
its viewing angle is different 
from the query image's viewing angle by $T^\theta$
or its viewing area $V(p^D)$ does not overlap with the query image's viewing area $V(p^Q)$, 
where $p^D$ and $p^Q$ are 
the ground-truth viewpoints of 
the destructor and the query images.
In this case, the viewing area is empirically defined as an isosceles-triangular region with an apex angle of 40 deg and the length of a leg as 50 m.

Localization performance is measured by its recognition rate.
Given a query image,
its retrieval result is in the form of 
a ranked list of database images (with length $N^D$).
Then,
the recognition rate $y$
is defined
over a set of self-localization tasks,
as 
the ratio $y$ 
of tasks
whose relevant database images
are correctly included in the top $x$
$(x\le N^D)$
ranked images.

\subsection{Localization Difficulty Index}\label{sec:ldi}

The core of the localization difficulty index (LDI) is the evaluation of
the view overlap between the relevant pair of query and database images.
Intuitively, the amount of view overlap can be evaluated by counting the
number of local features matched between the relevant pair. In this
study, we tested three different strategies for local feature matching:
SIFT matching without any post verification \cite{se2002mobile}, SIFT matching
with geometric verification by RANSAC \cite{lim2012real}, and using vector
field consensus (VFC) \cite{vfc}. We determined that VFC stably produces acceptable results. SIFT-matching frequently produces many false positives and is
not effective to identify image pairs with small overlap. RANSAC
geometric verification is effective only when there are many structured
objects such as buildings and does perform well in general cases.
Conversely, VFC is stable and able to produce many true
matches; it performs well in both structured and unstructured scenes.
Based on this result, we elected to implement VFC as the method for
evaluating view overlap in the following experiments.

Localization difficulty index $D(I^{query})$ 
is now defined as a decreasing function of view overlap $O(\cdot,\cdot)$ between query image $I^{query}$ and its relevant image $I^{relevant}$:
\begin{equation}
D(I^{query}) = 1 / \left[ O (I^{query}, I^{relevant}) \right]. \label{eqn:vo}
\end{equation}
Predicting localization difficulty
from such an image based cue is
an ill-posed problem;
it is impossible to
design a perfect prediction method.
Rather,
our strategy for difficulty prediction
is based on the
{\it relative}
LDI value.
More formally,
we sort all the $N^E$ self-localization tasks
in ascending order of LDI 
and then compare
the difficulty of different self-localization tasks
by using its rank within the sorted list
normalized by the list's length
[\%].
To create
a dataset,
we sample
pairs of query and its relevant database images
from a range of normalized rank 
$[rank^{min}, rank^{max}]$,
in which
the parameters
$rank^{min}$\% -  
$rank^{max}$\%,
control the relative difficulty of the dataset. 
In practice,
we observed
that this prediction method 
performed effectively.
Fig. \ref{fig:D} 
displays samples of 
viewpoints 
used in the experiment
for three distinct cases
corresponding to 
three different levels of LDI.

\figG

\section{Methods}

The proposed PCA-NBNN approach consists of three distinct steps: (1) modeling, (2) encoding and (3) retrieval of scenes, each of which is detailed in the following subsections.

\subsection{Modeling by Bag-of-Scene-Parts}\label{sec:partmodel}

Scene modeling is an important first step in visual robot self-localization. The objective of scene modeling is to convert a robot's view image to an invariant scene descriptor, which allows a robot to search over an environment map or a collection of pre-mapped view images to identify similar views. The main  problem we faced was how to describe a scene discriminatively and compactly, both of which are necessary to manage the geometric/photometric view changes and the significant amount of visual information. The proposed approach is inspired by the fact that even DCNN features frequently fail to capture the local parts of a scene, as we will see in the experimental discussion, Section \ref{sec:exp}. Typically, it is weak against large view changes and frequently produces poor results in visual robot localization. Hence, we adopt a kind of bag-of-parts scene model \cite{lcd5}, where each query/database image is described by an unordered collection of part-level features, to facilitate fast, yet discriminative correspondence between partially overlapping images. 

A key design issue is how to discover useful parts in a scene. This is
different from the problem of object segmentation, i.e., segmenting
an image into meaningful parts such as objects, which is a core
problem in the field of computer vision \cite{pandey2011scene}. Rather, our goal is to realize consistent segmentation for similar view images, allowing a
robot to obtain similar parts for similar scenes (i.e., relevant scene pair).  
In general, any part-segmentation technique such as clusters of superpixels described in our ICRA15 
paper \cite{kanji2015unsupervised}, can be adopted. 
In the current study, we borrow techniques
from the unsupervised object detector \cite{alexe2012measuring}, which quantifies how likely
it is for an image window to contain an object of any class. 
We first
extract the set of 100 bounding boxes with the highest objectness score
and then rerank these according to the area of the bounding box and
select the top ($N^P-1$) ranked parts. 
In total, we obtain a size $N^P=21$ set of DCNN features consisting of one image-level feature and 20 part-level features.

\subsection{Encoding by Experience-based Vocabulary}

We then encode the scene parts to a bag-of-parts representation \cite{lcd5}. First,
we extract a 4,096 dimensional DCNN feature from a region that
corresponds to the bounding box of each scene part. Although a DCNN
is composed of a number of layers, in each of which responses from
the previous layer are convoluted and activated by a differentiable
function, we use the sixth layer of DCNN, as it has proven to produce
effective features with excellent descriptive power in previous
studies \cite{tomomi2011incremental}. Then, we perform PCA compression to obtain 128
dimensional features. Our strategy is supported by the recent findings
in \cite{neuralcodes} where PCA compression provides excellent short codes
with 512, 256, and 128 short vectors that provide state-of-the-art accuracy on
a number of recognition tasks. In our experiments, we use DCNN
features from the database to train the PCA models for different
settings of the output dimension, 512, 256, and 128.

Another key design issue is an efficient scene retrieval. The 
bag-of-parts representation, presented above,
is a relatively compact and discriminative scene descriptor. 
However,
it is a high dimensional description and does not directly realize
high-speed scene retrieval. To address this issue, we adopt the nearest
neighbor approach \cite{kanji2015cross} where each local feature is explained by its
nearest neighbor (NN) library features. Because the original local feature
can be compactly represented by the IDs of NN library features, efficient
data structures such as inverted files can realize compact indexing and
fast retrieval.

One of the most popular instances of the NN approach is the bag-of-words 
(BoW) \cite{17}, a well-established technique for image retrieval.
Its key component is offline dictionary learning. That is, offline,
a set of visual features are extracted from training images and then
a dictionary of exemplar visual features are learned by unsupervised
learning algorithms such as k-means clustering. Once such a dictionary
is learned, a given image is translated to an unordered collection of
NN library features, each of which is compactly represented by the ID of
the NN library feature, which is termed visual word. This pre-learning
of the dictionary is effective to achieve fast retrieval. Conversely, a
known limitation of the BoW model is its vector quantization effect, which
significantly reduces the descriptive power of the BoW descriptor.

We address the above issue by an experience-based fine vocabulary. As
a key difference from the BoW approach, we directly employ a library of
available visual features, i.e., not the vector quantized version, termed
visual experience. This strategy is motivated by the fact that an enormous
amount of visual experience is readily available, such as a collection of
visual images acquired in the robot's previous navigation or shared
by colleague robots, as well as images crawled from the web. Because the proposed
approach does not rely on vector quantized visual features, database
features are expected to be approximated by many more similar NN
library features. For example, in \cite{kanji2015unsupervised}, we explored an approach for
common landmark discovery aiming at unsupervised discovery of
part-level library features that effectively explain a given input image.
In this study, we employ a simple nearest neighbor-based distance
metric to identify a library feature that approximates a given database
feature.

We define place class as a collection of NN library features
that approximates a given database image. To evaluate the dissimilarity
between a query and a place class (i.e., database image), we propose to employ
the image-to-class distance. This strategy is inspired by our previous
IROS15 paper \cite{kanji2015cross}, where image-to-class distance was successfully
applied to an alternative scenario of cross-domain localization using
SIFT features. Let $I$ and $C$ denote a given query image and a place
class (i.e., database image), both are represented by a set of local
features ($I=\{f\}$, $C=\{f'\}$),
then image-to-class distance is defined by:
\begin{equation}
f(I, C) = 
\sum_{f\in I}
\min_{f'\in C}
| f-f' |_2^2. \label{eqn:2}
\end{equation}

\figF

\subsection{Retrieval by Asymmetric Distance Computation}

The proposed scene retrieval strategy is an instance of asymmetric distance
computation (ADC), which only encodes the local features of the
database; not the query local feature (Fig \ref{fig:F}). 
This is in contrast to
symmetric distance computation (SDC) employed by typical BoW
systems, which encodes both query and database features. We have observed
that the SDC strategy performed poorly in the case of our enormous and
unorganized visual experience-based library owing to near duplicate and
useless library features. In fact, there is virtually no probability that a query
image and its relevant database image have the same NN library
features in common (Fig. \ref{fig:F} top). Hence, we encode only database
features and we directly match a query feature and each database
image's NN library features (Fig. \ref{fig:F} bottom).

ADC is more accurate than SDC and employed in some
previous systems in different contexts \cite{vlad}. 
In our view, ADC functions even when there are many near duplicate library
features, which is the case of our fine library. As another advantage,
ADC allows an incremental update (e.g., deletion/insertion of
features) of the database and the library, which is an important property
from the viewpoint of incremental mapping and localization \cite{ProbabilisticsRobotics}. 

However, ADC is computationally more demanding as it
requires many-to-many comparisons between the query and database images.
To address this issue, we employ a compact 
binary encoding of images
and fast bit-count operation that enables fast image comparison (Fig. \ref{fig:G}). 
Query and
library features are encoded to $N_1$ bit binary codes using the compact
projection technique borrowed from \cite{tomomi2011incremental} and \cite{min2010compact}
and compared by Hamming distance.
Another limitation of the original NBNN distance metric is that it must pre-define a set of place classes. 
To address this, 
as mentioned,
our algorithm mines the available visual experience (i.e., library) to locate similar $N^p$ NN library features that effectively explain the database feature, 
in the same spirit as in our previous IROS14 paper \cite{vem16}, and then use the set of mined $N^p$ library features as the place class that corresponds to the database image. 
Then, we compute the image-to-class distance between the query image to each place class by (\ref{eqn:2}).

\figI

\figU

\section{Experiments}\label{sec:exp}

We evaluated the performance over three independent datasets 
that were collected 
from different routes and viewpoints.
The datasets used in these experiments 
consisted of collections of view images 
captured around a university campus,
using the vision system described in \ref{sec:dataset}.
Fig. \ref{fig:L} presents an overhead view of our experimental environment and viewpoint paths.
For each viewpoint path,
we acquired a collection of dense view images. 
Occlusion is severe in
all the scenes and people and vehicles are dynamic entities occupying
the scenes. Moreover, viewpoints are close to each other, which
produces many near-duplicate database images and makes self-localization more difficult.

We investigated self-localization performance versus difficulty,
based on the performance difficulty index
introduced in \ref{sec:ldi}.
First, 
a number of samples of sets of 
query, relevant database image, and destructor database images
were generated,
and sorted
according in ascending order of the LDI
defined in (\ref{eqn:vo}).
Then,
five different sets of 
100 self-localization tasks
with different levels of difficulty
for each of the three viewpoint paths
were sampled
from rank
0\%-20\%,
0\%-50\%,
0\%-100\%,
50\%-100\%,
and 80\%-100\%
of the sorted list of self-localization tasks.
Note that our
strategy of down-sampling the original image set to a small (i.e., size 100) subset does
not sacrifice self-localization difficulty as long as we use the recognition
rate (defined in \ref{sec:lpi}) as the performance index.

\figE
\figH

In the experiments,
different versions of
image-level and part-level DCNN features were compared.
Image-level DCNN features include
the original 4,096-dim DCNN descriptor (``dcnn"),
its PCA compressed 128-dim, 256-dim and 512-dim descriptors (``pca128", ``pca256", and ``pca512"),
and 
``pca128"
descriptor
is further compressed by compact projection
to 20-bit, 16-bit and 12-bit code
(``bin20", ``bin16", and ``bin12").
For the sake of reproducibility,
we simply use 
the full libraries
with size $2^{20}$, $2^{16}$ and $2^{12}$
respectively
for 
the 20-bit, 16-bit and 12-bit codes.
Part-level DCNN features 
(``bodw20", ``bodw16", and ``bodw12")
are different from
``bin20", ``bin16", ``bin12"
only in that 
they are originated from
not only the image-level DCNN feature
but also from part-level DCNN features,
as described in \ref{sec:partmodel}.
Further,
we also implemented 
an alternative part-level feature,
termed ``bodf",
which is only different from the above
``bodw20/16/12"
in that 
the 128-dim PCA compressed part-level DCNN feature
was used without binarization.
We use the ``bodf" 
only for the purpose of investigating
the quantization loss 
caused by our compact projection.
Note that
in practice,
the ``bodf" 
method is not efficient and
requires 
relatively high
space and time cost.

Fig. \ref{fig:I}
is a summary of 
distributions of 
self-localization performance 
versus view overlap,
where the
self-localization performance 
is measured in terms of the normalized rank 
of the relevant database image
and
the view overlap is measured by VFC matches.

Fig. \ref{fig:U} 
shows 
samples of query and database image pairs
with view overlap score measured by VFC,
together with 
performance results from
12 different self-localization tasks
using the proposed ``bodw20" method.
The case \#4 
has relatively high ``overlap" value
due to false positive matches in the VFC verification,
despite the fact 
that it is one of hardest self-localization tasks 
and in fact its self-localization performance is bad, rank = 93 \%.
The case \#12
has 0 ``overlap" value
and despite the fact,
recognition algorithm performs relatively well, rank = 21 \%.
For several cases, such as \#1 and \#5, occlusion is a major source of errors in place recognition.
Despite the difficulty,
it can be said
that the proposed recognition algorithm
stably performs well
as will be shown in performance results (Figs. \ref{fig:E}, \ref{fig:H}).

Fig. \ref{fig:E} presents the
results for relatively easy
self-localization tasks.
Note that 
for the ``rank: 0\% - 20\%" dataset,
the proposed method
``bodw20"
with fine ($2^{20}$=) 1M vocabulary
performs relatively well
despite the fact
it is much more efficient than 
non-binarized 
DCNN features.
Its rank-10\% identification rate
is approximately 0.9
and comparable to that of 
non-binarized DCNN features
``dcnn" and 
high dimensional features
``pca256" and ``pca512".
Unfortunately,
the performance of the proposed method
becomes relatively less than the 
non-binarized
high-dimensional
DCNN features
for self-localization with medium level difficulty,
as indicated in ``rank: 0\% - 50\%" and 
``rank: 0\% - 100\%".
This indicates relative robustness 
of non-binarized DCNN feature 
in self-localization with a medium level of difficulty.

Fig. \ref{fig:H} presents the
results for relatively hard
self-localization tasks.
It can be seen 
that the proposed method ``bodw20"
again produces comparable results to 4,096-dim DCNN features.
Its top-10\% identification rate
is comparable to that of 
non-compressed or PCA-compressed versions
of DCNN features 
``dcnn", ``pca128", ``pca256", ``pca512", and ``bodf".
This is because the fact
that 
in the relatively hard self-localization scenarios,
the performance of DCNN features
drop drastically,
because the query scene appears quite different from the relevant 
database scene.
It can be said
that the proposed ``bodw" method
achieved
good tradeoff between
efficiency and descriptive power
in the challenging scenario of
self-localization from images with small overlap.

Finally,
we investigated 
the case of
``self-localization from images with NO overlap".
More formally,
we considered a hardest setting
where the relevant database image with nearest neighbor viewpoint
had zero overlap in terms of 
the number of matches by VFC.
Fig. \ref{fig:H}``NO" displays
the results.
It can be seen that 
DCNN features
are better than chance (i.e., $y=x/300$).
There are two reasons for this: 
(1) VFC may fail to detect matches even when there is view overlap between the relevant image pair;
(2) views of relevant image pairs are frequently similar to one another owing to the atmosphere effect even when there is no view overlap. 
Overall, 
the proposed method ``bodw20" is comparable to the non-binarized
DCNN features, despite the fact 
that it is computationally significantly
more efficient.

\section{Conclusions}

In this paper, we addressed the problem of self-localization from
images with small overlap. We explicitly introduced a localization
difficulty index as a decreasing function of view overlap between
query and relevant database images and investigated performance
versus difficulty for 
challenging cross-view self-localization tasks. 
We then presented a
novel approach to bag-of-visual-features scene retrieval called PCA-NBNN 
to facilitate fast, yet discriminative correspondence 
between partially overlapping images. In experiments, we investigated
localization performance versus difficulty and confirmed that the proposed
method frequently yielded comparable performance with 
non-binarized high-dimensional DCNN features.
We further addressed an alternative important scenario of
``self-localization from images with NO overlap", 
where the highly compressed PCA-NBNN feature is comparable to the previous high-dimensional DCNN features.

\bibliographystyle{IEEEtran}
\bibliography{cvl}

\end{document}